\documentclass[a4paper]{article}
\usepackage{INTERSPEECH2019}
\usepackage{times}
\usepackage{helvet}
\usepackage{courier}
\usepackage{times}
\usepackage{latexsym}
\usepackage{graphicx}
\usepackage{url}
\usepackage{tipa} 
\usepackage{verbatim}
\usepackage{multirow}
\usepackage{algorithm,algorithmic}

\title{PhonSenticNet: A Cognitive Approach to Microtext Normalization for Concept-Level Sentiment Analysis}
\name{Ranjan Satapathy$^1$, Aalind Singh$^2$, Erik Cambria$^1$}
\address{
  $^1$SCSE, Nanyang Technological University\\
  $^2$Vellore Institute of Technology, India
  }
\email{satapathy.ranjan@ntu.edu.sg, singh.aalind@gmail.com, \\cambria@ntu.edu.sg}
\begin{document}
\maketitle
\begin{abstract}
  With the current upsurge in the usage of social media platforms, the trend of using short text (microtext) in place of standard words has seen a significant rise. The usage of microtext poses a considerable performance issue in concept-level sentiment analysis, since models are trained on standard words. This paper discusses the impact of coupling sub-symbolic (phonetics) with symbolic (machine learning) Artificial Intelligence to transform the out-of-vocabulary concepts into their standard in-vocabulary form. The phonetic distance is calculated using the Sorensen similarity algorithm. The phonetically similar in-vocabulary concepts thus obtained are then used to compute the correct polarity value, which was previously being miscalculated because of the presence of microtext. Our proposed framework increases the accuracy of polarity detection by 6\% as compared to the earlier model. This also validates the fact that microtext normalization is a necessary pre-requisite for the sentiment analysis task.
\end{abstract}


\section{Introduction}
\label{intro}
With the popularization of mobile phones and Internet social networks, the use of electronic text messaging, or texting, has reached astonishing figures such as more than 8,000 tweets produced per second\footnote{\url{http://www.internetlivestats.com/one-second/}}. These type of communications are usually performed in real time and over platforms which impose limitations on the length of the messages, as in the case of Twitter or the traditional SMS system. Because of this, the writing style of these messages differs from normal standards and phenomena such as word shortenings, contractions and abbreviations are commonly used both to gain writing speed and circumvent length limitations. Moreover, even in the case of messaging platforms where length restrictions do not apply (e.g. WhatsApp), it is also common to see a writing style which tries to better reflect the feelings of the writer. 
Given that most data today is mined from the web, microtext analysis is vital for many natural language processing (NLP) tasks. In the context of sentiment analysis, microtext normalization is a necessary step for pre-processing text before polarity detection is performed~\cite{camsui}. 

The two main features of microtext are relaxed spelling and reliance on emoticons and out-of-vocabulary (OOV) words involving phonetic substitutions (e.g., `b4' for `before'), emotional emphasis (e.g., `goooooood' for `good') and popular acronyms (e.g., `otw' for `on the way')~\cite{Read2005,Rosa2009,Xue2011}. 
The challenge arises when trying to automatically rectify and replace them with the correct in-vocabulary (IV) words~\cite{Liu2011}. It could be thought that microtext normalization is as simple as performing find-and-replace pre-processing~\cite{Khoury2015}. However, the wide-ranging diversity of spellings makes this solution impractical (e.g., the spelling of the word ``tomorrow'' is generally written as ``tomorow, 2moro, tmr among others). Furthermore, given the productivity of users, novel forms which are not bound to orthographic norms in spelling can emerge. For instance, a sampling of Twitter studied in~\cite{Liu2011} found over $4$ million OOV words where new spellings were created constantly, both voluntarily and accidentally. 
Concept-based approaches to sentiment analysis focus on a semantic analysis of text through the use of web ontologies or
semantic networks, which allow the aggregation of conceptual and affective information associated with natural language
opinions. The analysis at concept-level is intended to infer the semantic and affective information associated with natural language opinions and hence, to enable a comparative fine-grained feature based sentiment analysis. In this work, we propose PhonSenticNet, a concept based lexicon which advantages from phonetic features to normalize the OOV concepts to IV concepts. The  International Phonetic Alphabet (IPA)\footnote{\url{https://www.internationalphoneticassociation.org/content/full-ipa-chart}} is used as the phonetic feature in the proposed framework.

The rest of the paper is organised as follows: Section \ref{sec:related_works} discusses the literature survey in microtext, Section \ref{sec:datasets} discusses the datasets used, Section \ref{sec:experiments} discusses the experiments performed and Section \ref{sec:conc} concludes the work done with future directions for this work.
\section{Related Work}
\label{sec:related_works}
This section discusses the work done in the domain of microtext analysis.
 \begin{figure*}[ht!]
\centering
\includegraphics[width = \textwidth]{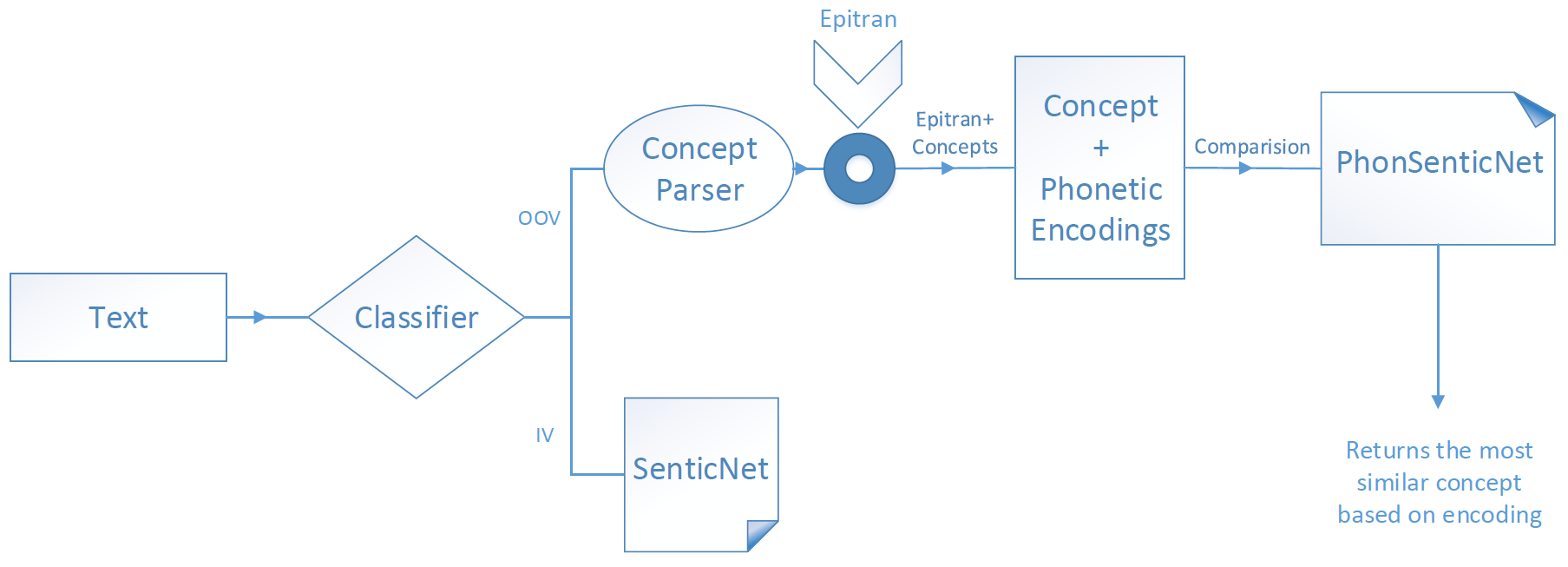}
\caption{Architecture of the framework}
\label{fig:arch}
\end{figure*}
\subsection{Microtext Analysis}
Microtext has become ubiquitous in today's communication. This is partly a consequence of Zipf's law, or principle of least effort (for which people tend to minimize energy cost at both individual and collective levels when communicating with one another), and it poses new challenges for NLP tools which are usually designed for well-written text~\cite{Hutto2014}. Normalization is the task of transforming unconventional words or concepts to their respective standard counterpart. ~\cite{satapathyphonetic} uses Soundex algorithm to transform out-of-vocabulary to in-vocabulary and shows it's effect on sentiment analysis task.



In~\cite{li08}, authors present a novel unsupervised method to translate Chinese abbreviations. It automatically extracts the relation between a full-form phrase and its abbreviation from monolingual corpora, and induces translation entries for the abbreviation by using its full-form as a bridge. ~\cite{han2011lexical} uses a classifier to detect OOV words, and generates correction candidates based on morpho-phonemic similarity. 
The types and features of microtext are reliant on the nature of the technological support that makes them possible. This means that microtext will vary as new communication technologies emerge. In our related work, we categorized normalization into three well-known NLP tasks, namely: spelling correction, statistical machine translation (SMT), and automatic speech recognition (ASR).
\subsubsection{Spelling Correction}
Correction is executed on a word-per-word basis which is also seen as a spell checking task. This model gained extensive  attention in the past and a diversity of correction practices have been endorsed by \cite{Church1991,Brill2000,Li2006,Pennell2011,Toutanova2002}. Instead,~\cite{Sproat2001} and \cite{Cook2009} proposed a categorization of abbreviation, stylistic variation, prefix-clipping, which was then used to estimate the probability of occurrence of the characters. Thus far, the spell corrector became widely popular in the context of SMS, where \cite{ChoudhuryM2007} advanced the hidden Markov model whose topology takes into account both ``graphemic'' variants (e.g., typos, omissions of repeated letters, etc.) and ``phonemic'' variants. All of the above, however, only focused on the normalization of words without considering their context.
\subsubsection{Statistical Machine Translation}
When compared to the previous task, this method appears to be rather straightforward and better since it has the possibility to model (context-dependent) one-to-many relationships which were out-of-reach previously~\cite{Kobus2008}. Some examples of works include \cite{Aw2006,Kaufmann2010,Pennell2014}. However, the SMT still overlooks some features of the task, particularly the fact that lexical creativity verified in social media messages is barely captured in a stationary sentence board. 

\subsubsection{Automatic Speech Recognition}
ASR considers that microtext tends to be a closer approximation of the word's phonemic representation rather than its standard spelling. As follows, the key to microtext normalization becomes very similar to speech recognition which consists of decoding a word sequence in a (weighted) phonetic framework.
For example, \cite{Kobus2008} proposed to handle normalization based on the observation that text messages present a lot of phonetic spellings, while more recently~\cite{Khoury2015} proposed an algorithm to determine the probable pronunciation of English words based on their spelling. Although the computation of a phonemic representation of the message is extremely valuable, it does not solve entirely all the microtext normalization challenges (e.g., acronyms and misspellings do not resemble their respective IV words' phonemic representation). Authors in~\cite{Beaufort2010} have merged the advantages of SMT and the spelling corrector model.

\section{Datasets}
\label{sec:datasets}
This section introduces to datasets used. The twitter dataset is available on request. The concept-level lexicon SenticNet is publically available \footnote{\url{http://sentic.net/senticnet-5.0.zip}}. 

\subsection{NUS SMS Corpus}
This corpus 
has been created from the NUS English SMS
corpus\footnote{\url{http://github.com/kite1988/nus-sms-corpus}}, wherein ~\cite{wang2013beam} randomly selected 2,000 messages. The messages were
first normalized into standard English and then translated into standard Chinese. 
For our training and testing purposes, we only used the actual messages and their 
normalized English version. It also contains non-English terms, which LSTM had no 
problem in learning. Singlish is an English-based creole that is lexically and 
syntactically influenced by Hokkien, Cantonese, Mandarin, Malay and 
Tamil~\cite{brown1999singapore}. It is primarily a spoken variety in Singapore, to 
emerge as a means of online communication for 
Singaporeans~\cite{warschauer2002internet}. 

\subsection{SenticNet}
SenticNet~\cite{cambria2018senticnet} is a knowledge base  of 100,000 commonsense concepts. Sentic API provides the semantics and sentics (i.e., the denotative and connotative information) associated with the concepts of SenticNet 5, a semantic network of commonsense knowledge that contains 100,000 nodes (words and multiword expressions) and thousands of connections (relationships between nodes). We used concept parser~\cite{poria2014dependency} in order to break sentences to concepts and analyze them. The concepts in the SenticNet contains their corresponding polarities.

\begin{table}[ht!]
\centering
\scriptsize

\begin{tabular}{|c|c|c|c|}
\hline
\textbf{Concepts} & \textbf{Polarity} &\textbf{Soundex Encoding} & \textbf{IPA Encoding} \\\hline
a\_little &	Negative& A000\_L340 & \textipa{\ae\_lIt\ae l} \\ \hline
abandon 	& Negative &A153 & \textipa{\ae b@nd\ae n} \\ \hline
absolutely\_fantastic 	& Positive &A124\_F532& \textipa{@bs@lutlI\_f@nt@stIk} \\ \hline

\end{tabular}

\caption{Sample Soundex and IPA Encodings with polarities for SenticNet5}
\label{tab:soundex with senticnet}
\end{table}
\subsection{Normalized Tweets}
The authors in~\cite{satapathyphonetic}, built a dataset which consists of tweets and their transformed in-vocabulary counterparts. We demonstrate our results by extracting concepts from unconventionally written sentences and then passing them through our proposed module to convert them to standard format concepts' and their corresponding polarities from SenticNet.

\begin{table*}[ht!]
\centering
\scriptsize
\caption{Precision, Recall, F1 and Accuracy for each algorithm on different datasets}
\begin{tabular}{|l|c|c|c|c|c|c|c|c||c|c|c|c|c|c|c|c|}
\hline
\multirow{3}{*}{} & \multicolumn{8}{c||}{\textbf{NUS SMS Dataset}}                                                                                                                                                                                              & \multicolumn{8}{c|}{\textbf{Twitter Dataset}}                                                                                                                                                                                              \\ \cline{2-17} 
                  & \multicolumn{2}{c|}{\begin{tabular}[c]{@{}c@{}}Logistic -\\ Regression\end{tabular}} & \multicolumn{2}{c|}{SGDC}    & \multicolumn{2}{c|}{SVC} & \multicolumn{2}{c||}{\begin{tabular}[c]{@{}c@{}}Multinomial-\\ NB\end{tabular}} & \multicolumn{2}{c|}{\begin{tabular}[c]{@{}c@{}}Logistic -\\ Regression\end{tabular}} & \multicolumn{2}{c|}{SGDC}   & \multicolumn{2}{c|}{SVC}  & \multicolumn{2}{c|}{\begin{tabular}[c]{@{}c@{}}Multinomial-\\ NB\end{tabular}} \\ \cline{2-17} 
                  & IV                                        & OOV                                      & IV            & OOV          & IV           & OOV         & IV                                     & OOV                                   & IV                                        & OOV                                      & IV           & OOV          & IV            & OOV         & IV                                     & OOV                                   \\ \hline
Precision         & 0.91                                      & 0.95                                     & 0.84          & 0.98         & 0.87         & 0.97        & 0.89                                   & 0.97                                  & 0.71                                      & 0.69                                     & 0.63         & 0.72         & 0.74          & 0.67        & 0.81                                   & 0.68                                  \\ \hline
Recall            & 0.95                                      & 0.90                                     & 0.99          & 0.81         & 0.98         & 0.85        & 0.97                                   & 0.87                                  & 0.68                                      & 0.71                                     & 0.80         & 0.52         & 0.64          & 0.77        & 0.61                                   & 0.85                                  \\ \hline
F-measure         & 0.93                                      & 0.92                                     & 0.91          & 0.89         & 0.92         & 0.91        & 0.93                                   & 0.92                                  & 0.69                                      & 0.70                                     & 0.70         & 0.60         & 0.68          & 0.72        & 0.69                                   & 0.76                                  \\ \hline
Accuracy          & \multicolumn{2}{c|}{\textbf{0.9275}}                                                          & \multicolumn{2}{c|}{0.89875} & \multicolumn{2}{c|}{0.915} & \multicolumn{2}{c||}{0.9225}                                                    & \multicolumn{2}{c|}{0.6962}                                                          & \multicolumn{2}{c|}{0.6605} & \multicolumn{2}{c|}{0.7013} & \multicolumn{2}{c|}{\textbf{0.7288}}                                                    \\ \hline
\end{tabular}
	
\label{tab:perf}
\end{table*}

\section{Experiments}
\label{sec:experiments}
This section dives into experiments performed to develop a concept-level microtext normalization module. The experiments performed help in deciding the best set of parameters to achieve state of the art accuracy. We name the lexicon which we built from SenticNet as PhonSenticNet. It contains concepts and their related phonetic encoding which is extracted from Epitran~\cite{Epitran}. 

\subsection{Framework}
The architecture of the proposed model is depicted in Figure \ref{fig:arch}. The framework classifies a sentence as OOV or IV using binary classifier. Following this, the OOV sentence is passed through the concept parser, and then the concepts are transformed to IPA using Epitran. The IPA of OOV concepts are matched to the PhonSenticNet, and then the IV concept is fetched. The corresponding polarity of the IV concept is retrieved from SenticNet. The detailed procedure is explained in the following subsections:
\subsubsection{Classification of microtext}
In this subsection, we employ various binary classifiers to detect microtext so as to reduce the execution time of the overall algorithm. We observed that the execution time of polarity detection task was reduced by 20\%. Different classifiers were trained on the two datasets namely NUS SMS data and Twitter dataset as shown in Table \ref{tab:perf}.

We use the term frequency–inverse document frequency (TF-IDF)~\cite{ramos2003using} approach for the task of feature extraction from a given text. We first split the document into tokens assigning weights to them based on the frequency with which it shows up in the document along with how recurrent that term occurs in the entire corpora. We used this approach to train four different classifiers. The evaluation metrics such as Precision, Recall, F-measure and Accuracy have been enlisted in the Table \ref{tab:perf}.

\subsubsection{Soundex vs IPA}
We compare our proposed IPA based method to Soundex~\cite{satapathyphonetic} since only~\cite{satapathyphonetic} have incorporated phonetic features to improve sentiment analysis. Soundex gives a lot of duplicate encoding
, whereas IPA gives no duplicate encoding. Each concept is unique in phonetic subspace, thereby increasing the efficiency for microtext normalization at concept-level. 
The number of concepts present in the lexicon is $100000$. The duplicates\footnote{Repetition of a soundex encoding for greater than one} due to Soundex encoding are $46080$. This shows that using soundex for microtext normalization has some information loss at the concept-level. As a result of Soundex encoding, we have $46080$ ambiguous concepts which affect the microtext normalization in real time. Hence, we propose to use IPA for all the phonetic based microtext normalization methods. The IPA based encoding has no redundancy, and thereby no information loss occurs during microtext normalization. 

\begin{figure}
    \centering
    \includegraphics[width = 0.51\textwidth]{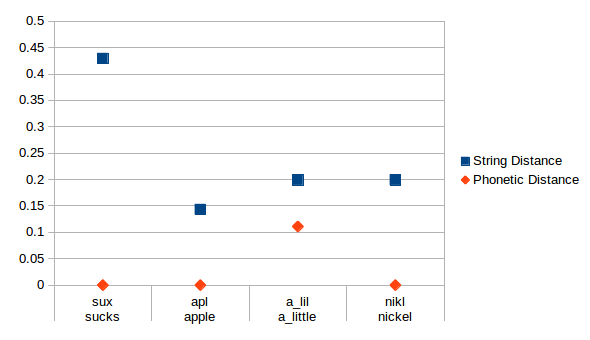}
    \caption{Visualization of string and phonetic distance}
    \label{fig:dist}
\end{figure}

\begin{algorithm}
	\caption{Algorithm for microtext normalization using phonetic features}
	\label{phon}
	\begin{algorithmic}[]
		\STATE Sentence (S) $=$ $s_{1}, s_{2}$,$\cdots$, $s_{n}$
		\STATE $c_{i}$ = concept-parser(S)
		\STATE \textbf{For} each concept $c_{i}$ in $S{n}$
		\STATE closest-match-concept = PhonSenticNet($c_{i}$)
		\IF{closest\_ match($C_{i}$, SenticNet)}
		\STATE  return concept polarity
		\ELSE
		\STATE  return sentence polarity
		\ENDIF
		\STATE average over polarity of concepts for sentence polarity
		\textbf{EndFor}
		\RETURN{} sentence polarity
	\end{algorithmic}
\end{algorithm}

\begin{algorithm}
	\caption{Closest Match Algorithm}
	\label{close}
	\begin{algorithmic}[]
		\STATE Concept (C) $=$ $c_{1}, c_{2}$,$\cdots$, $c_{m}$
		\STATE \textbf{For} each concept $c_{i}$ in C
		\STATE Sorensen ($c_{i}$,Senticnet)
		\STATE \textbf{EndFor}
		\STATE \textbf{return} phonetically closest matching concept
	\end{algorithmic}
\end{algorithm}

\subsubsection{Microtext normalization of concepts using IPA}
The analysis at concept-level is intended to infer the semantic and affective information associated with natural language opinions and, hence, to enable a comparative fine-grained feature-based sentiment analysis. Concept-based approaches to sentiment analysis focus on a semantic analysis of text through the use of web ontologies or semantic networks, which allow the aggregation of conceptual and affective information associated with natural language opinions. In order to normalize concepts found on the social media, we built a resource for concept-level phonetic encodings by using concepts from SenticNet. We used concept parser~\cite{poria2014dependency} to extract concepts from the input text. The concepts are then transformed to a subspace where they are represented by their phonetic encodings. Table \ref{tab:soundex with senticnet} shows sample concepts with their respective soundex encodings and IPA from SenticNet 5.

Phonetic encoding transforms concepts from the string subspace into their phonetic subspace. This transformation eliminates the redundant concept encoding produced by Soundex. The input concept is then passed through this phonetic subspace (IPA used in PhonSenticnet) to find the most phonetically similar concept and then returns it. The algorithm \ref{phon} and \ref{close} describe the procedures in detail.

\begin{figure}[h!]
	\centering
	\includegraphics[width = 0.48\textwidth]{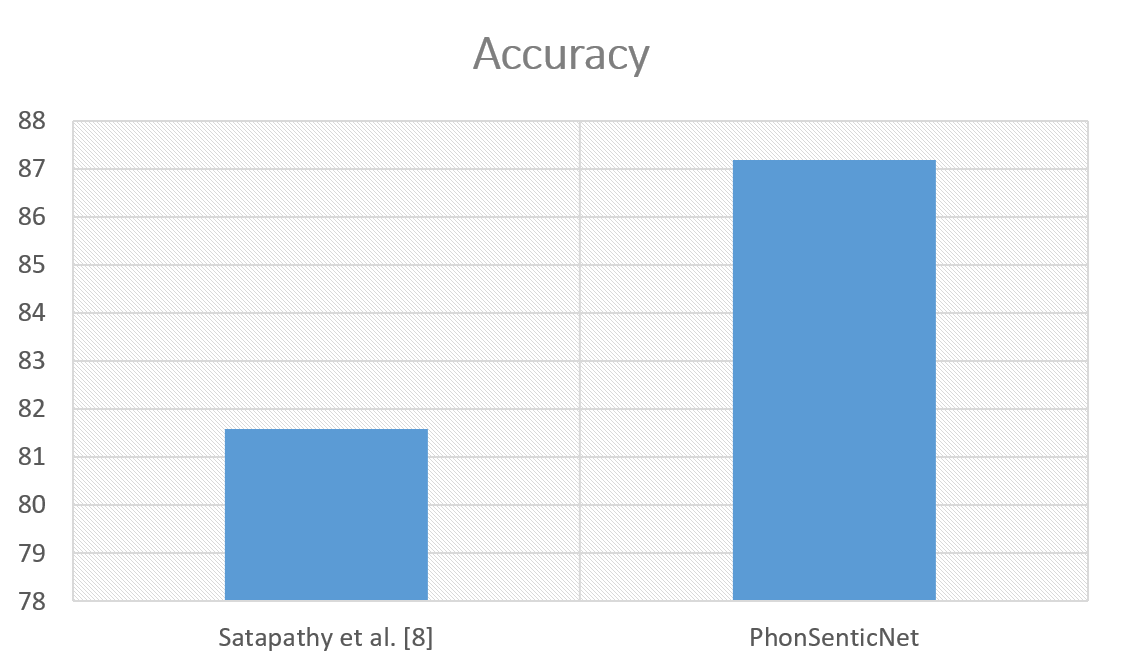}
	\caption{Accuracy for polarity detection}
	\label{fig:acc}
\end{figure}
\subsubsection{Polarity Detection with SenticNet}
While SenticNet 5 can be used as any other sentiment lexicon, e.g., concept matching or bag-of-concepts model, the
right way to use the knowledge base for the task of polarity detection is in conjunction with sentic patterns, sentiment-specific linguistic patterns that infer polarity by allowing affective information to flow from concept to
concept based on the dependency relation between clauses. The sentiment sources of such affective information are extracted from SenticNet 5 by firstly generalizing multiword expressions and words by means of conceptual primitives and, secondly, by extracting their polarity.
We compare our polarity detection module results with~\cite{satapathyphonetic}. The accuracy increases significantly by 6\% as shown in Figure \ref{fig:acc}.

\begin{table}[h!]
\centering
\scriptsize
\begin{tabular}{|l|c|c|}
\hline
\multicolumn{1}{|c|}{\textbf{Text}} & \textbf{\begin{tabular}[c]{@{}c@{}}Sentence Polarity before \\microtext normalization\end{tabular}} & \textbf{\begin{tabular}[c]{@{}c@{}}Sentence Polarity  after \\microtext normalization\end{tabular}} \\ \hline
I wil kil u                       & Neutral                                                                              & Negative                                                                                \\ \hline
m so hapy                           & Neutral                                                                              & Positive                                                                                \\ \hline
i dnt lyk reading                  & Positive                                                                             & Negative                                                                                \\ \hline
it is awesum 2 ride byk           & Neutral                                                                              & Positive                                                                                \\ \hline
\end{tabular}
\caption{Sample sentences before and after microtext normalization}
\label{fig:pol}
\end{table}


\section{Discussion and Future Work}
\label{sec:conc}
The proposed resource contains concepts from SenticNet and their phonetics by using Epitran which we name as PhonSenticNet. This resource is used as a lexicon for microtext normalization. The input sentence is broken down into concepts and then transformed into their phonetic encoding. The phonetic encoding is matched with the PhonSenticNet, the resource built in this work. Then the most similar matching concept and it's the corresponding polarity is returned as shown in the algorithm \ref{phon} and \ref{close}. 
\begin{enumerate}
	\item We have taken Sorensen similarity to measure the distance. The Sorensen similarity shows how similar the two input texts are to one another, where 0 means similar and 1 means dissimilar as shown in Figure \ref{fig:dist}.
	\item Figure \ref{fig:dist} shows some of the distance metric between non-standard and their standard concepts. The similarity is shown at both string and phonetic level.
	\item Previous paper~\cite{satapathyphonetic} shows sentence-level sentiment analysis, whereas in this work we focus on concept-level microtext normalization.
	\item It can be observed from Table~\ref{tab:perf} that the twitter dataset does not perform as good as the NUS SMS data. The reason behind it is, the twitter dataset contains acronyms like lol, rofl, etc instead of phonetic substitution. This also suggests how the way of writing differs in both messages and tweets.
\end{enumerate}

Microtext is very much language-dependent: the same set of characters could have completely different meaning in different languages, e.g., `555' is negative in Chinese language because the number `5' is pronounced as `wu' and `wuwuwu' resembles a crying sound but positive in Thai since the number 5 is pronounced as `ha' and three consecutive 5s correspond to the expression`hahaha'. Hence, we are working on it's multilingual version \cite{vilares2018babelsenticnet}. The proposed work only works for the phonetic class of microtext analysis. Though, the acronyms still rely on the lexicon built in~\cite{satapathyphonetic}. 


\bibliographystyle{IEEEtran}

\bibliography{sentic,microtext,naaclhlt2018} 


\end{document}